# SYNERGY OF ALL-PURPOSE STATIC SOLVER AND TEMPORAL REASONING TOOLS IN DYNAMIC INTEGRATED EXPERT SYSTEMS

Galina V. Rybina (*galina@ailab.mephi.ru*)
Alexey V. Mozgachev (*avmozgachev@mephi.ru*)
Dmitry V. Demidov (*dvdemidov@mephi.ru*)

National Research Nuclear University MEPhI (Moscow Engineering Physics Institute)
Kashirskoye shosse 31, Moscow, 115409, Russian Federation

*The paper discusses scientific and technological problems of dynamic integrated expert systems development. Extensions of problem-oriented methodology for dynamic integrated expert systems development are considered. Attention is paid to the temporal knowledge representation and processing.*

## INTRODUCTION

Methods of artificial intelligence are applied today in many industrial and socially significant areas. Qualitatively new opportunities of intelligent systems (in particular, dynamic intelligent systems) allow to increase efficiency of automation in known areas and to solve harder problems for which traditional methods and software aren't suited [Rybina, 2008].

Impact of using intelligent systems depends on availability of software tools and variety of their applications in strategically significant areas. This leads to intense researches and software development in such areas as dynamic integrated expert systems, intelligent agents, multiagent systems and makes important the development of software tools and libraries that support dynamic intelligent systems implementation.

Despite the lack of semantically unified notion base and classification of dynamic intelligent systems the scope of scientific and technological problems interfering wide application of dynamic intelligent systems is already defined:

1. Difficulties of acquisition of knowledge that considers time from various sources such as experts, texts, databases, etc.
2. Difficulties of knowledge representations development for dynamic domains which can be characterized by evolving number of entities and external data updating in time.
3. The need for threading temporal reasoning for multiple asynchronous processes; high performance requirements for real-time applications; evolving data and even knowledge at runtime.
4. Problems of modeling the dynamic system environment and its real-time evolution patterns.
5. High costs of special development tools for dynamic intelligent systems construction.
6. The need for special hardware for interacting with outside world (sensors, controllers, etc.).

The problems above substantially define the development complexity of dynamic intelligent systems and, in particular, dynamic integrated expert systems which are mostly demanded. Yet there is no universal approach addressing all these problems. Even modern software tools like Gensym G2, RTworks, RTXPS, etc. aren't capable to solve them all.

New extensions of problem-oriented methodology offered and experimentally investigated by G. V. Rybina in recent years may become a considerable step towards the needed methodology for dynamic intelligent systems development. On the basis of problem-oriented methodology there was developed a unique software tool for knowledge engineers – the AT-TECHNOLOGY workbench for applied integrated expert systems construction. It constitutes from an empty integrated expert system with scalable architecture and a wide range of implemented models and methods for solving both ill-formalized and formalized problems.

Analysis shows that approaches for acquisition, representation, and processing of temporal knowledge are least studied [Rybina, 2008]. Ability to represent and use temporal dependences between occurring events allows to considerably reduce search space and affects the performance of dynamic integrated expert system. In modern tools (Gensym G2, RTworks, RTXPS, etc.) representations of time and temporal knowledge are rather simple and actually stay unused in the course of inference [Popov et al, 1996], [Rybina et al, 2010].

Thus, there is a need for models, methods, and software tools that deal with temporal reasoning in dynamic integrated expert systems. These models, methods, and software have to be integrated within uniform methodology and technology such as problem-oriented methodology together with AT-TECHNOLOGY workbench.

## BASIC MODELS AND METHODS OF REASONING IN STATIC INTEGRATED EXPERT SYSTEMS

Model of inference utilized for static problem domains in the AT-TECHNOLOGY workbench considers both reliable and unreliable knowledge [Rybina, etc., 2007]. Such negative factors as uncertainty, inaccuracy, fuzziness, subdefiniteness of data and knowledge, and some others (together called NEG-factors) are taken into account. The key feature of this model is that all calculations with NEG-factors including fuzzification and defuzzification are carried out as required during inference.

On each inference loop partial evaluations of rule antecedents are performed. To maximize the use of evaluation results they are stored in working memory for the further loops. Truth-values of logical expressions lie within the set $[0;\ 1] \cup NE$ where NE stands for «not evaluated yet». NE corresponds to a case when a truth-value of an antecedent can't be calculated because working memory lacks some facts used in antecedent.

In matching phase there may be a need for transformations of NEG-factors for joint processing of rules containing unreliable information. A number of fuzzification methods are implemented for each data type and NEG-factor. Our experience in building integrated ES tells that fuzzy knowledge makes 1-2% of knowledge base. Therefore using only fuzzy inference for such knowledge isn't adequate. So we do not totally fuzzify all data during inference but we fuzzify values only needed to calculate formulas with fuzzy values. At the end of inference we defuzzify all the values that became fuzzy. Membership functions are used to represent fuzzy values. It should be noted that during transformations in the course of inference some fuzzified values may become multimodal. In such situations our defuzzyfication method may give a set of accurate values corresponding to one multimodal membership function.

To accelerate conflict resolution we use a multicriteria algorithm of conflict resolution where calculations of rule ranks are done at matching phase. We consider specificity of rules, novelty of the facts, reliability of consequents. The conflict set is always ordered by rule ranks and persists through inference loops.

At backward chaining with lack of input data there are used some heuristics to ask for the needed parameter values:
1. First ask for parameters that are most frequently used in rules to give more chances to antecedents become fully evaluated.
2. First ask for parameters with larger value domains because they are possibly used in bigger number of competing hypotheses.
3. First ask for parameters that can be found in left-most conjunctions of LHS. Note that all conjunctions in rule antecedents are equally transformed from infix form like "a & b & c & d" where elements are ordered by importance to a prefix tree of the form "& (a, & (b, & (c, d)))".
4. First ask for parameters that can be found as mutually exclusive evidences in rules of the form "$a_i$ & B $\rightarrow$ $H_m$" and "~$a_i$ & C $\rightarrow$ $H_n$", where $H_m$, $H_n$ - hypotheses, B = $b_1$& … &$b_n$, C = $c_1$& … &$c_n$ – conjuncts, $a_i$ – weighed evidence. Getting one evidence excluded the other one and solver can reject one of the rules.

In all-purpose solver there are implemented control rules for unreliable knowledge processing. Processing of uncertainty, inaccuracy, subdefiniteness, and fuzziness separately can be fulfilled with the coefficients, attributes, intervals and membership functions assigned to values, statements, expressions and rules. But to process all that together we need to consider each pair of factors and design special matching techniques and methods of evaluating heterogeneous formulas. So at the end of inference any parameter value can bear in addition uncertainty or inaccuracy.

For interpretation of LHS and RHS of rules in all-purpose solver there were implemented arithmetic, logical, and comparison operations for various types of operands and NEG-factors. Basic arithmetic is extended to consider uncertainty and inaccuracy for the cases when at least one operand value is uncertain or inaccurate. If at least one operand is a membership function then we use fuzzification methods, fuzzy logic and Zadeh's extension principle. For linguistic variables which values are linguistic terms (strings) all-purpose solver replaces the terms by corresponding membership functions and also applies Zadeh's extension principle. Truth-values are all converted to [0; 1] and may stay not evaluated as well.

**MODELS AND METHODS OF TEMPORAL REASONING IN DYNAMIC INTEGRATED EXPERT SYSTEMS**

Analysis of different models and methods of representation of time and temporal dependences takes important place in modern researches of dynamic intelligent systems. They are described in [Shoham, 1987], [Shoham et al. 1988], [Allen, 1991], [Spranger, 2002], [Yeremeyev et al, 2004], [Stefanyk, 2007]. Some works are devoted to dynamic control [Moore, 1993], [Osipov, 2008], [Osipov, 2011], to point linear model of time [Yeremeyev et al, 2004], [Yeremeyev et al, 2009], to interval temporal logics [Allen, 1983], [Plesnevich, 1999], [Yeremeyev et al, 2012], to branching time [Ladkin et al, 1990], [Yeremeyev, 2006]. In [Rybina et al, 2010], [Rybina et al, 2013] the use of Allen's interval logic [Allen, 1983] and control in time [Osipov, 2008] is justified for the dynamic extension of the AT-TECHNOLOGY workbench.

This paper focuses on the model of temporal reasoning on production rules for dynamic integrated expert systems (ES) (Fig. 1). The offered model provides processing of knowledge containing temporal dependences together with basic knowledge in problem domain. Inference tools implemented within the framework of problem-oriented methodology [Rybina et al, 2007] [Rybina et al, 2013] are used.

Temporal reasoning considers temporal and cause-effect relations between objects in the domain of interest. It also considers the possibility of changes both in data and knowledge. Cyclic operation is the key feature of temporal solver on production rules in dynamic integrated ES. It means that input data may be updated each cycle and inference upon these data must be fulfilled each cycle as well. Fig.1 shows the described cycle. Its model is a generalization of the basic inference model described in [Rybina et al, 2007] [Rybina, 2008] for temporal issues.

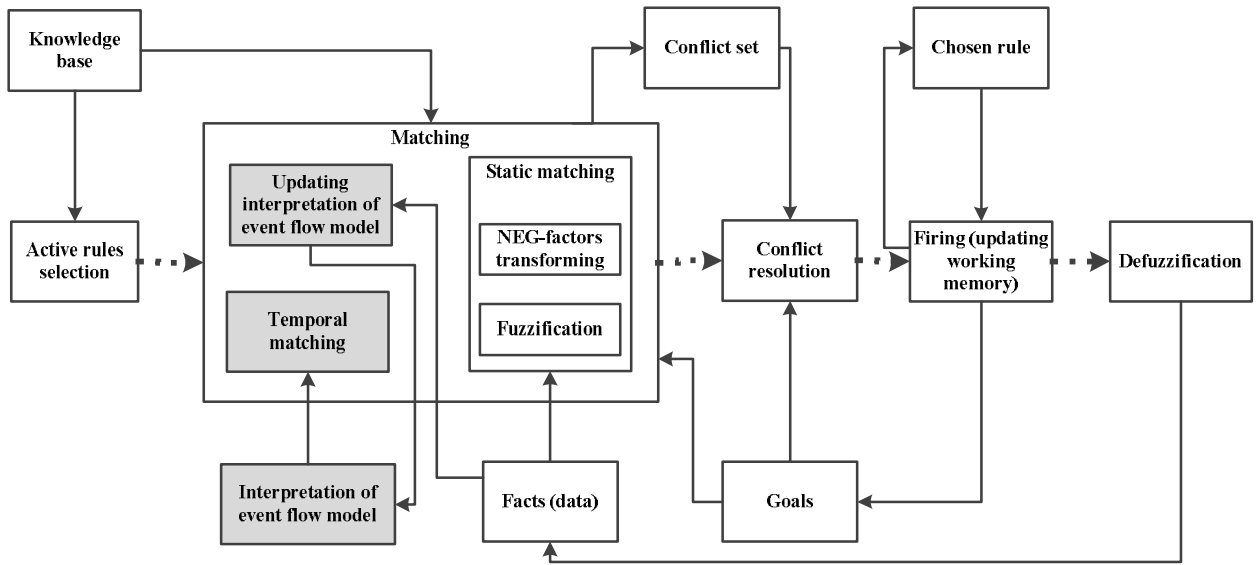

Fig. 1. Cycle of temporal reasoning on production rules

Formally this model of temporal inference can be given as follows:
I' = <A, S'', K, W', D>, where

A – active rules selection from knowledge base (KB);

S'' – group of matching and special activities including matching left-hand side (LHS) of active rules with working memory data; interpreting event flow model; matching temporal and static fragments of rules; transformations and conversions of such negative factors as uncertainty, inaccuracy, fuzziness, subdefiniteness, and some others (together called NEG-factors); fuzzification.

K – conflict resolution – choosing single rule from the conflict set to fire;

W' – rule firing – interpreting right-hand side (RHS) of the chosen rule considering NEG-factors;

D – defuzzification of data fuzzified during inference.

In Fig.1 activities that support temporal reasoning are shown in gray. They extend the model of inference with unreliable knowledge described in [Rybina et al, 2007] and [Rybina, 2008]. It is necessary to mention that in dynamic integrated ES inference is carried out independently on each operation cycle, event flow model persists through cycles, but its interpretation may change. Major challenges in integrating basic and temporal inference tools are: implementation of efficient reinterpreting event flow model; implementation of matching of global interpretation of event flow model built for all types of temporal expressions with local event flow models in production rules.

Thus, there are two goals of temporal reasoning: interprete current event flow model and generate the list of control actions to affect problem domain (or its model).

For representing problem domains in dynamics there were chosen Allen's logic [Allen, 1983] with some modifications and logic of control in time [Osipov, 2008].

As it was shown in [Rybina et al, 2013], Allen's classical logic lacks some important features such as momentary events, quantitative characteristics of temporal primitives, etc. That complicates its application in dynamic integrated ES. So Allen's logic was modified to support these facilities based on ideas described in [Plesnevich, 1999].

We shall provide here the description of Allen's logic with modifications. The main symbols are:
- variables designating events and intervals;
- variable attributes: .c – number of origins of event or interval, .l – duration;
- logical relations: ~ (negation), & (conjunction), v (disjunction);
- Allen's connectives: b (before), a (after), m (meets), o (overlaps), s (starts), d (during), e (is equal, synchronous with), f (finishes);
- comparison operations: > (greater), < (less), = (equal);
- integers;
- left and right parenthesis: ( and ).

Formulas of Allen's logic with modifications are:
- X r Y expressions where X, Y – intervals and r – any Allen's connective;
- X r Y expressions where X, Y – events and r – one of {b, a, e};
- X r Y expressions where X – an event, Y – an interval, r – one of {b,a,s,d,f};
- XH r N expressions where X – a variable, H – its attribute, r – a comparison operation, N – an integer.
- variables;
- simple formulas;
- ~f, (f & g), (f v g) where f and g are formulas.

Event flow model is specified by a set of temporal objects (events and intervals). Local event flow model in a rule is specified by Allen's logic formulas. Interpretation of event flow model is affected by the event occurrence times and by the intervals' start and end times. Application of Allen's logic with modifications assumes the concept of discrete time, i.e. changes in problem domain happen within cycles only.

Mechanisms of control in time allow to apply certain actions upcoming from changes in state of objects without introducing time explicitly ([Moore, 1993], [Osipov, 2008], [Osipov, 2011]). Applying these mechanisms implies implementation of different methods of rule activation allowing to react to specific observed events and intervals.

For temporal knowledge representation in dynamic integrated ES there was extended basic knowledge representation language (KRL) of the AT-TECHNOLOGY workbench [Rybina, 2008]. The KRL allows to represent basic problem domain knowledge including knowledge with uncertainty, inaccuracy and fuzziness. Generalized KRL in addition allows to represent temporal knowledge based on Allen's interval logic with modifications and logic of control in time. For this purpose there were introduced :
- new object types (event and interval),
- new types of attributes (logical expressions for origin conditions),
- changing in structure of LHS for the sake of local event flow model considerations,
- new types of rules (conventional, periodic, response).

We shall now describe the method of temporal knowledge processing. It allows to perform reasoning on knowledge containing temporal dependences by matching global event flow model interpretation with local event flow models. The main changes in basic inference model are incapsulated in matching phase:
- interpreting event flow model;
- processing temporal fragments of LHS of production rules.

Interpretation of event flow model puts events and intervals on time axis by identifying origins based on working memory data and reviewing history of their origins in the past. Special attention is given to cases of repeated origins and non-standard layout of events on time axis. For example, termination of an interval before its opening. Processing of temporal fragments of LHS of production rules uses the results of global event flow model interpretation. Local event flow models of active rules are checked to comply with this interpretation by means of Allen's logic with modifications and logic of control in time.

## TEMPORAL INFERENCE TOOLS IN DYNAMIC INTEGRATED EXPERT SYSTEMS

As a result of the research work there was developed temporal reasoning tools (temporal solver) deeply integrated with the all-purpose solver within the AT-TECHNOLOGY workbench [Rybina, 2008]. Temporal solver is included into dynamic extension of the AT-TECHNOLOGY workbench. That allows dynamic integrated ES successfully operate both in static and dynamic problem domains.

During operation temporal solver performs two basic activities: interpreting event flow model and instantiating temporal fragments of LHS of production rules. Functional requirements to temporal reasoning tools stated in [Rybina et al, 2014]. Fig. 2 shows the architecture of temporal solver.

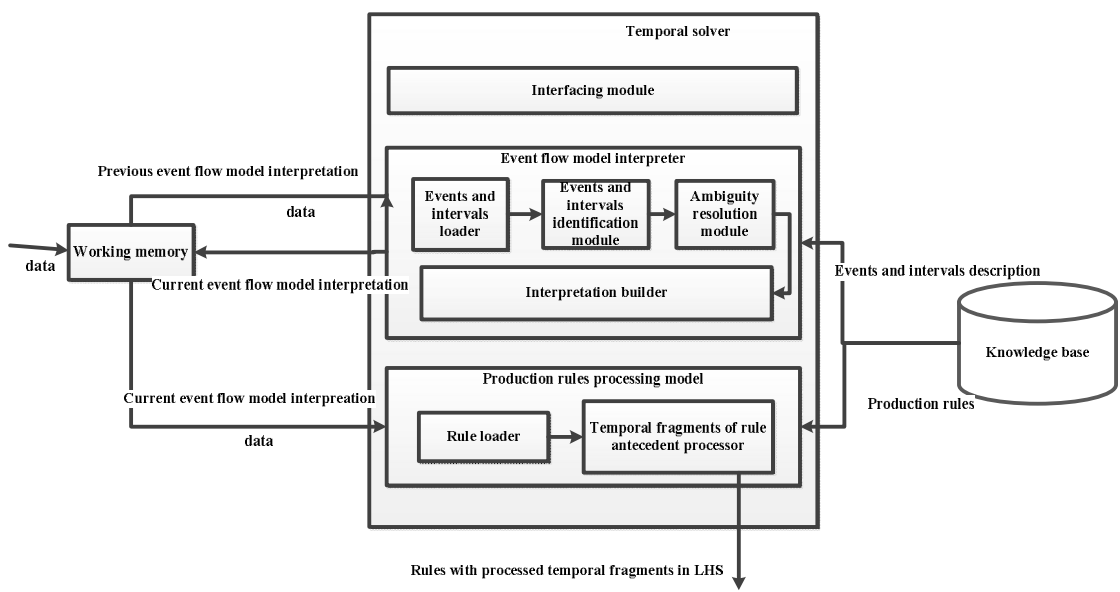

Fig. 2. Temporal solver architecture

Besides temporal solver there was developed some facilitation tools for collaborative functioning of components of expert systems and simulation modeling software within the new architecture of dynamic integrated ES. Fig. 3 shows the diagram of interaction of temporal solver, all-purpose static solver, and simulation modeling tools. Some components also use working memory and dynamic blackboard to exchange data.

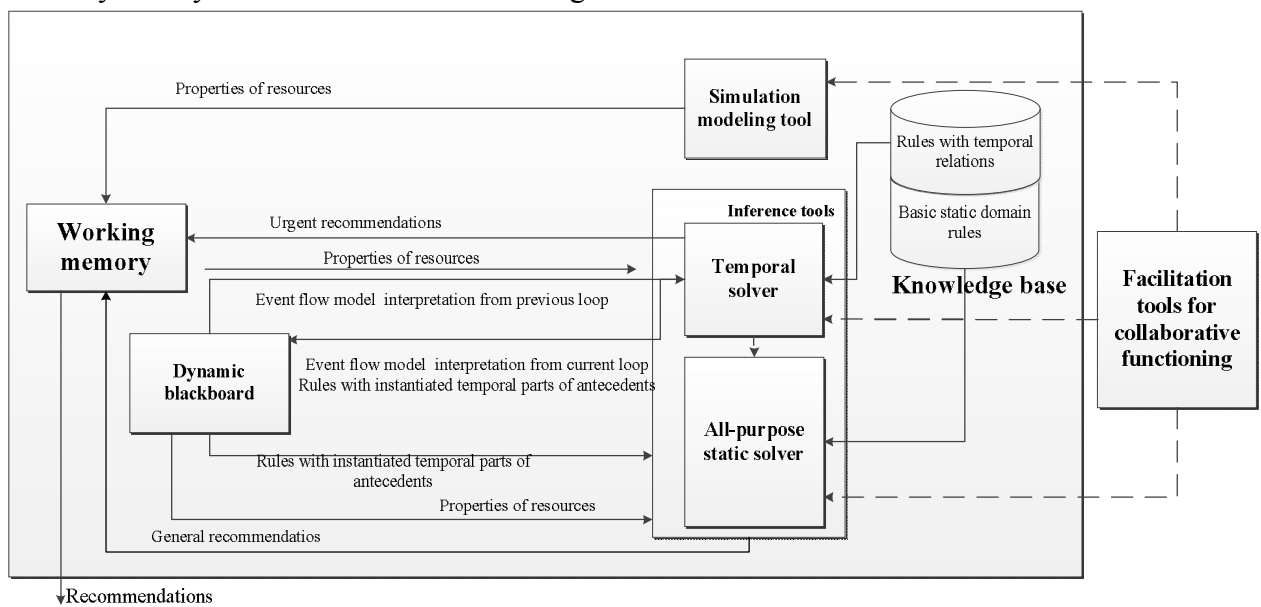

Fig. 3. Interaction diagram of temporal solver, all-purpose static solver, and simulation modeling tool.

## CONCLUSION

Recent experiments showed some advantages of the developed software over competitors by the following criteria: KRL power, speed, dynamic integrated ES construction time. There are under construction several prototypes of dynamic integrated ES of big practical importance.

## ACKNOWLEDGEMENTS

This work was partially funded by the RFBR (Russian Foundation for Basic Research), project No. 12-01-00467.